# Towards Real-Time Visual Tracking with Graded Color-names Features

Lin Li[1][0000-0001-9897-6041] Xuemei Guo*[0000-0001-5298-8063] Guoli Wang[1][0000-0003-4031-7996]

[1],* Sun Yat-sen University, Guangzhou, Gaungdong Province, China
guoxuem@mail.sysu.edu.cn

**Abstract.** MeanShift algorithm has been widely used in tracking tasks because of its simplicity and efficiency. However, the traditional MeanShift algorithm needs to label the initial region of the target, which reduces the applicability of the algorithm. Furthermore, it is only applicable to the scene with a large overlap rate between the target area and the candidate area. Therefore, when the target speed is fast, the target scale change, shape deformation or the target occlusion occurs, the tracking performance will be deteriorated. In this paper, we address the challenges above-mentioned by developing a tracking method that combines the background models and the graded features of color-names under the MeanShift framework. This method significantly improve performance in the above scenarios. In addition, it facilitates the balance between detection accuracy and detection speed. Experimental results demonstrate the validation of the proposed method.

**Keywords:** Visual Tracking, MeanShift,.

## 1 Introduction

With the continuous development of computer technology, the popularization of Internet and edge computing, artificial intelligence is developing towards the deep level of social application. Moving object detection and tracking technologies based on video image are critical to autonomous robot applications. Its purpose is to determine the position, speed, scale and other attributes of moving object in each frame of an image sequence according to the spatio-temporal correlation. In last two decades, a large number of mature algorithms have been applied to specific environments and achieved great results [1, 2]. However, due to the influence of dynamic factors in real life [3, 4, 5], such as cloud shadow, leaf swing, illumination variations, scale variations and occlusions, the detection and tracking of moving target is still a very challenging subject. In addition, heterogeneous cyber-physical systems [6] put forward higher requirements for the execution time, state reachability, memory usage and latency of the computing model.

In this paper, we improve the traditional MS in three steps. Initially, according to the differences of the image gray values and the spatio-temporal correlation of these differences [7], we find out area of the moving candidate target, so as to establish the background models. Secondly, the image is segmented by using the color-names [8] features of it. Base on the above, the information entropy of color-names is used as the weight of Meanshift vector to estimate the position and speed of the target. Finally, the confidence degree of the target is calculated. If the value is too low, it is considered that occlusion or deformation has occurred. If so, the compound gradient method is used to perform graded matching on the components of the candidate target. In this way, we make MS better adapt to the above complex environment. Furthermore, the simplicity of the method proposed here makes it easy to be extended or trimmed to different computing capabilities.

The remainder of this paper is organized as follows. Section 2 reviews the related work on visual tracking. The detailed method, i.e. background modeling, extraction of color-names feature and graded matching process, are described in Section 3. Section 4 describes the experimental evaluation and analysis. Section 5 summary and future work outlook.

## 2  Related Work

Generative methods[10, 11, 12, 13] and discriminative methods [9] are two different categories in target tracking.

### 2.1  Generative Methods

Common generative tracker, such as mean shift tracker[10, 11, 12], cam tracker, and etc., extracts target features and uses the generated models to match. This kind of methods are usually efficient but can not deal with the problems, such as illumination variations, shadows, and etc..

Danelljan et al. proposed the color-names features as adaptive color attributes, which changes the input features into 11 color labels, and then uses the idea of PCA to transform the feature dimension into a two-dimensional retraining correlation filter. This kind of method is very efficient and can adapt to illumination variations. But it model and update the overall appearance of the target, so it is not good at the problems of target occlusions, scaling, non rigid motion change, and etc..

To solve the above problems, some visual trackers have adopted the method of multi filter [1] or multi-scale spatial pyramid [14, 15], which estimates the optimal scale by sampling each component and multi-scale of the moving target, and then repeatedly applying the correlation filter. This kind of methods have achieved great success in detecting target scale changes, and improve the problem that the information decays exponentially with time caused by the linear combination of the template and the previous frame template. However, it will bring higher time and space complexity.

In addition, all these generative methods need to use prior knowledge to model the tracking target, which is difficult to apply to the unknown environment. At this point, discriminative methods usually have advantages.

## 2.2 Discriminative Methods

Discriminative tracker [16, 17] realizes tracking through the difference between the target models and the background information, and transforms the tracking problem into a binary classification problem to find out the decision boundary between the moving target and the background. This makes prior knowledge no longer necessary, but also brings other problems. When they encounter disturbances by the clutter of the background, they will cause the background models to update continuously.

Codebook [3], VIBE [5] and kernel density prediction method (KDE) all are classical background modeling algorithms. The basic idea behind them is that in the long-term observation, the background accounts for most of the time, and more data support the background distribution. At the same time, in order to deal with the multi-mode and periodic change characteristics of the background in real scene, such as swinging leaves and periodically refreshing billboards, they have established time series models for the pixels in the image. However, these algorithms almost establish sequence models for each pixel without difference, resulting in the increase of time and space complexity when the number of pixels is large.

# 3 Visual Tracking with Graded Color-names Features

## 3.1 Background Modeling

Like classical background modeling methods mentiond above, the background modeling in this paper is also based on Gaussian theory. The difference lies in what kind of spatio-temporal sequence models are established to cope with the multimode nature of the background. We only established spatio-temporal sequence models for moving blocks rather than for each pixel in an image. In this section, we will briefly introduce the method of background modeling based on object pre-extraction.

Initially, as shown in Eq.(1, 2), calculate the initial value of expectation and variance of RGB value for each pixel in the image. When time $t = N$, we have

$$\mu_t(x, y) = \frac{1}{N} \sum_{t=0}^{N-1} l_t(x, y) \tag{1}$$

$$\sigma_t^2(x, y) = \frac{1}{N} \sum_{t=0}^{N-1} (l_t(x, y) - \mu_t(x, y))^2 \tag{2}$$

where $\mu$ is expectation, $\sigma^2$ is variance and $l$ is RGB value of current image.

Secondly, gray and normalized the current image. This step can suppress the interference of light intensity change effectively. Again, compare it with the gray values of the background models which get from the first step. Determine whether the difference between the two gray values of each pixel exceeds 2 times of the standard deviation. From the definition of the standard deviation of a normal distribution follows that statistically about 5% of the sampled pixels will be excluded. If so, they will be regarded as moving points. So far, there are still a large number of false positive noise points in the detection results, and the moving target is prone to be "holes". Finally, to improve this problem, the gray histogram is used to further the pre-extraction process.

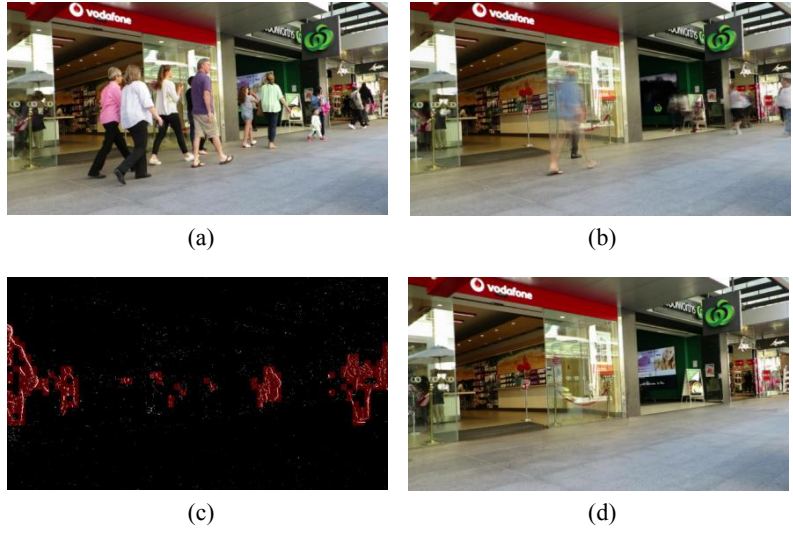

(a)                          (b)

(c)                          (d)

**Fig. 1.** In the MOT16 image sequence (a), because there are a large number of people in the image, and there are fast-moving targets and slow-moving targets at the same time. Gaussian background modeling is always difficult to update or frequently jitter (b). In this paper, the method of object pre-extraction is adopted to screen out the blocks (c) belonging to the moving target. Then, different learning rates apply to different cases, foreground or background (d).

Note that we are based on two assumptions: Assumption 1: Pixels will not move in isolation. Assumption 2: The moving object has similar appearance in an image sequence, that is, there is a similar gray density sequence in some region of the image sequence. As a result, a block is regarded as a moving block only when the points in it judged as moving exceeds a threshold (e.g. 10%); Otherwise, the block is regarded as noise. Then, the whole image is divided into small pieces of equal width and height, hereinafter referred to as blocks.

$$F_i = \frac{n_i \cdot (h - h_\varepsilon)}{n \cdot h} \quad (3)$$

$F_i$ is the gray density, $n_i$ represents the number of the moving points in the block, $n$ represents the total number of the points in the block, $h_\varepsilon$ is the distance from the block to the center of the block group. $h$ is the radius of the block group. As shown in Fig.2.

Here, in consecutive frames, the blocks moving in the same direction, at the similar speed and are adjacent to each other are classified into one group and considered to be a candidate target.

$$\mu_t = (1 - \alpha) \cdot \mu_{t-1} + \alpha \cdot l \quad (4)$$

$$\sigma_t^2 = (1 - \alpha) \cdot \sigma_{t-1}^2 + \alpha \cdot (l - \mu_t)^2 \quad (5)$$

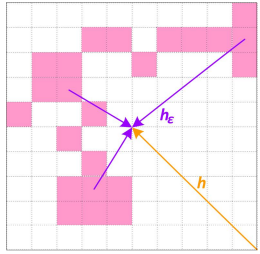

**Fig. 2.** The group of moving block

where $\alpha$ is the learning rate. When update the expectation and variance of the background, the learning rate of pixels in the scope of moving blocks is low, and vice versa. Fig.1 shows that in the above complex case, the proposed method can work well.

### 3.2 Color-names Features

In the process of detection and tracking, most trackers adopt density or texture features [18, 19, 20, 21], however, color is invariant in the process of image translation, rotation and scaling, so choosing color as the basis of image matching has strong robustness. MS makes use of the difference of the color histogram to track and locate the target. Meanshift vector is shown in Eq.(6).

$$m(x) = \left[ \frac{\sum_{i=1}^{n} g\left(\left\|\frac{x-x_i}{h}\right\|^2\right) \cdot \omega(x_i) \cdot (x_i - x)}{\sum_{i=1}^{n} g\left(\left\|\frac{x-x_i}{h}\right\|^2\right) \cdot \omega(x_i)} \right] \quad (6)$$

$x$ is the starting point of the search. $x_i$ is the pixel location in the current frame, $h$ is the kernel bandwidth, $g(\cdot)$ is kernel function, $\omega(\cdot)$ is weight function. $m(x)$ is the desired Meanshift vector in one iteration.

Note that there are two important differences between the Meanshift vector in this paper and that of the traditional MS: the starting point and the weight function.

On the one hand, the starting point in the traditional MS is the target center in the previous frame, while in this paper we search from the center of the block group in current frame, that is, the center of the candidate target, which we get from the process of pre-extraction. On the other hand, in most of the literature [12] using MS, the weight of a pixel is inversely proportional to the distance between the pixel and the target center, so as to eliminate the error caused by the target edge noise. However, the internal color of a target usually tends to converge, and the edge weight is low, which easily leads to the Meanshift vector offset. That is, contour information is very important for target matching.

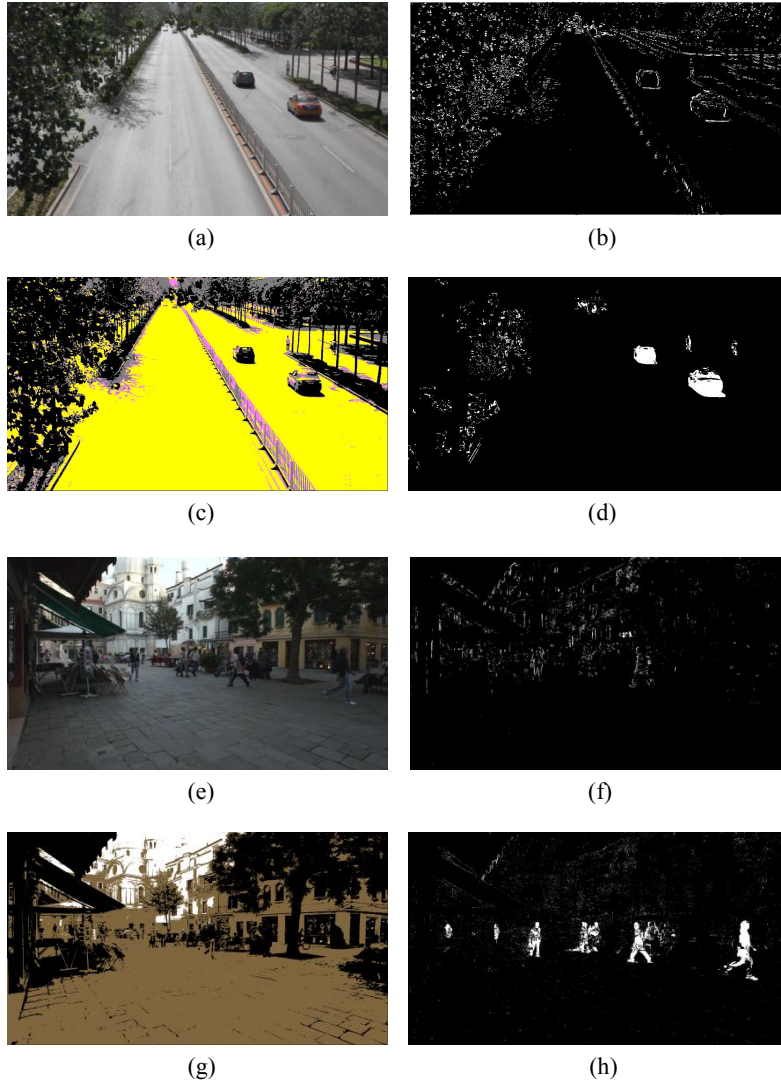

Fig. 3. In the MVI/MOT image sequence (a, e), due to the interference of noise, when extracting moving pixels based on frame difference method (b, f), the moving pixels are completely submerged in the noise. In this paper, background modeling, image segmenting (c, g) and target detection are processes of mutual influence and restriction. Based on these, it can accurately judge the pixels that really belong to the moving target (d, h).

In order to make more effective use of the contour information, it is necessary to make the pixels be stably classified in an image sequence. Now, we will briefly analyze the commonly used color spaces in MS. RGB color space is widely used in MS, but it is hardware oriented. The three components of RGB color space are closely related to brightness, that is, as long as the brightness changes, the three components will change accordingly. And it is not intuitive to continuously change of color. HSV color space is widely used in image processing [22, 23, 24]. H and S are usually be used to segment and detect images. But it still belongs to high-dimensional features, which leads to an increase in computational complexity. What is more troublesome is that one pixel is easily misclassified due to image noise. In 2014, Danelljan et al. proposed to apply color-names [8] to extending the tracker and achieved remarkable results. Color-names are based on the basic classification of colors in the human world, that is, RGB colors are subdivided into 11 types: black, blue, brown, gray, green, orange, pink, purple, red, white and yellow. In the field of computer vision, color-naming is an operation that associates RGB observations with linguistic color labels. We intend to use color-names as the label of color to generate feature template.

The dimension of features should not be too low to maintain the identifiability, and it should not be too high, resulting in wrong classification of pixels in an image sequence. It requires that the distance of inter-class is as large as possible, while that of intra-class is as small as possible. To meet the above requirements, we use MMSE to reduce the number of color labels.

$$J_s(a) = \|Ya - b\|^2 = \sum_{i=1}^{n}(a^t y_i - b_i)^2 \tag{7}$$

$$\nabla J_s = \sum_{i=1}^{n} 2(a^t y_i - b_i)y_i = 2Y^t(Ya - b) \tag{8}$$

$$S_w = \sum_{i=1}^{2} \sum_{x \in D_i}(x - m_i)(x - m_i)^t \tag{9}$$

$$a = \alpha n S_w^{-1}(m_1 - m_2) \tag{10}$$

In Eq.(7, 8), $Y$ is the sample matrix, $b$ is the type of the samples, $a$ represents a projection transformation, $J_s(\cdot)$ is the mean square error function. To minimize $J_s(\cdot)$, $\nabla J_s$ must be equal to zero, as shown in Eq.(8). In Eq.(9), $m_i$ represents the mean value of type $i$. $S_w$ is the total scatter matrix, it represents the distance of intra-class. It is easy to prove that, as shown in Eq.(10), the best mapping transformation meets our needs.

That is, MMSE can be used to select the most representative color labels for us. First, the number of color labels is manually specified. The image is projected to the color labels, then the weight of each color label is calculated according to Eq.(11). Those labels with the largest weight are selected as the bases.

$$\omega_{label} = \min(\nabla H) \cdot \frac{n_i}{n} \tag{11}$$

$n_i$ is the number of pixels belonging to the color label $i$, $n$ represents the total number of pixels, and $\nabla H$ is the minimum difference between the current color label and other color labels have been choosed.

Then, the information entropy of color-names is used as the weight of Meanshift vector. According to the information entropy theory: The higher the probability of occurrence, the lower the amount of information it carries, and vice versa. When a natural image is regarded as a random process, its color distribution conforms to the Gaussian distribution, that is, the fewer pixels belonging to a certain color label in an image, the higher the identifiability of this label. The weight of Meanshift vector is calculated as follows:

$$\omega(x) = -C \cdot \sum_{x \in X_j} p(x) \cdot \log p(x) \tag{12}$$

$$p(x) = n_j / n \tag{13}$$

$C$ is a normalized parameter, $n_j$ is the number of pixels belonging to the color label $j$.

### 3.3 Graded Features

Inspired by the event triggered controller used in [15, 25] control engineering, after the above steps, we will measure the confidence degree of the target. It is calculated as follows:

$$D = \sum_{i=sx}^{ex} \sum_{j=sy}^{ey} \left[ \frac{(C_{i,j} == P_{i+\Delta i, j+\Delta j} ? 1 : 0) \cdot \omega_{i,j}}{\omega} \right] \tag{14}$$

In Eq.(14), $D$ is the confidence degree, $(sx, sy)$, $(ex, ey)$ is the start and end position of the target in previous frame, $(\Delta x, \Delta y)$ is the displacement of the target in the current frame, $\omega_{i,j}$ is the weight of the pixel $(i, j)$, and $\omega$ is a normalized parameter.

When the confidence degree is lower than a threshold, it is most likely that serious occlusion or deformation has occurred. Then, segment the target templates according to the moving blocks that get in the object pre-extraction stage, and do matching gradly according to the local feature templates of the target. The valid range of position, direction and speed of the target in current frame are predicted according to the information of them in previous frames. For example, the estimated value of the target displacement is $(\Delta x, \Delta y)$, then the limit range of it is shown as Eq.(15, 16).

$$\Delta x_{valid} : (\lambda_{\min} \times \Delta x \sim \lambda_{\max} \times \Delta x) \tag{15}$$

$$\Delta y_{valid} : (\lambda_{\min} \times \Delta y \sim \lambda_{\max} \times \Delta y) \tag{16}$$

The parameter values in this paper are as follows: $\lambda_{min} = 0.5, \lambda_{max} = 2.0$.

The limit range are regarded as the constraint conditions of target matching. Then, matching process can be regarded as the optimization problem of unimodal function with constraints. We start searching from a certain point in the feasible region. At the beginning, it iterates along the gradient direction. Once it leaves the feasible region, it changes direction so that the search process is always carried out in the feasible region. As shown in Fig.4.

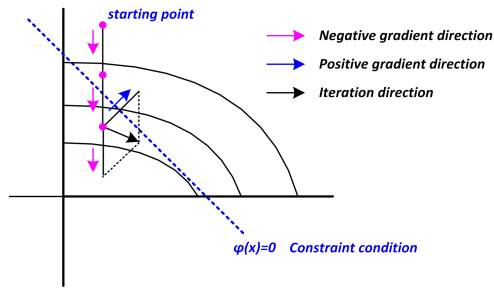

**Fig. 4.** Calculate iteration direction

## 4      EXPERIMENTS

### 4.1    Accuracy

Fig.5 shows the tracking result of the MVI image sequence. Observe the tracking effect under the conditions of rapid vehicle movement, scale variations, occlusions between people and vehicles, overlap between people. The traditional MS takes the 45th frame as the starting frame, Fig.5(a~f) shows that it is difficult to deal with the problems above. As shown in Fig.5(g~r), we takes the first frame as the starting frame, after the improvement proposed in this paper, it can achieve good robustness to problems above in long-term tracking.

### 4.2    Performance

On an Intel(R) Core(TM) i7-4600M, 2.9GHz machine, it can process 27 frames(640× 480 pixels) per second.

## 5      CONCLUSION

This paper presents a vision based target detection and tracking method. It requires no prior knowledge of background or the target. Experimental results showed that the proposed method outperforms the traditional MS under the conditions of scale variations, occlusions, shape deformations, and etc., achieved stable effect in long-term tracking. At the same time, the improvement of accuracy does not significantly reduce the tracking efficiency. This method can still achieve a processing speed of 27 frames(640×480 pixels) per second, which is close to the upper limit of human eye resolution of 30 frames per second.

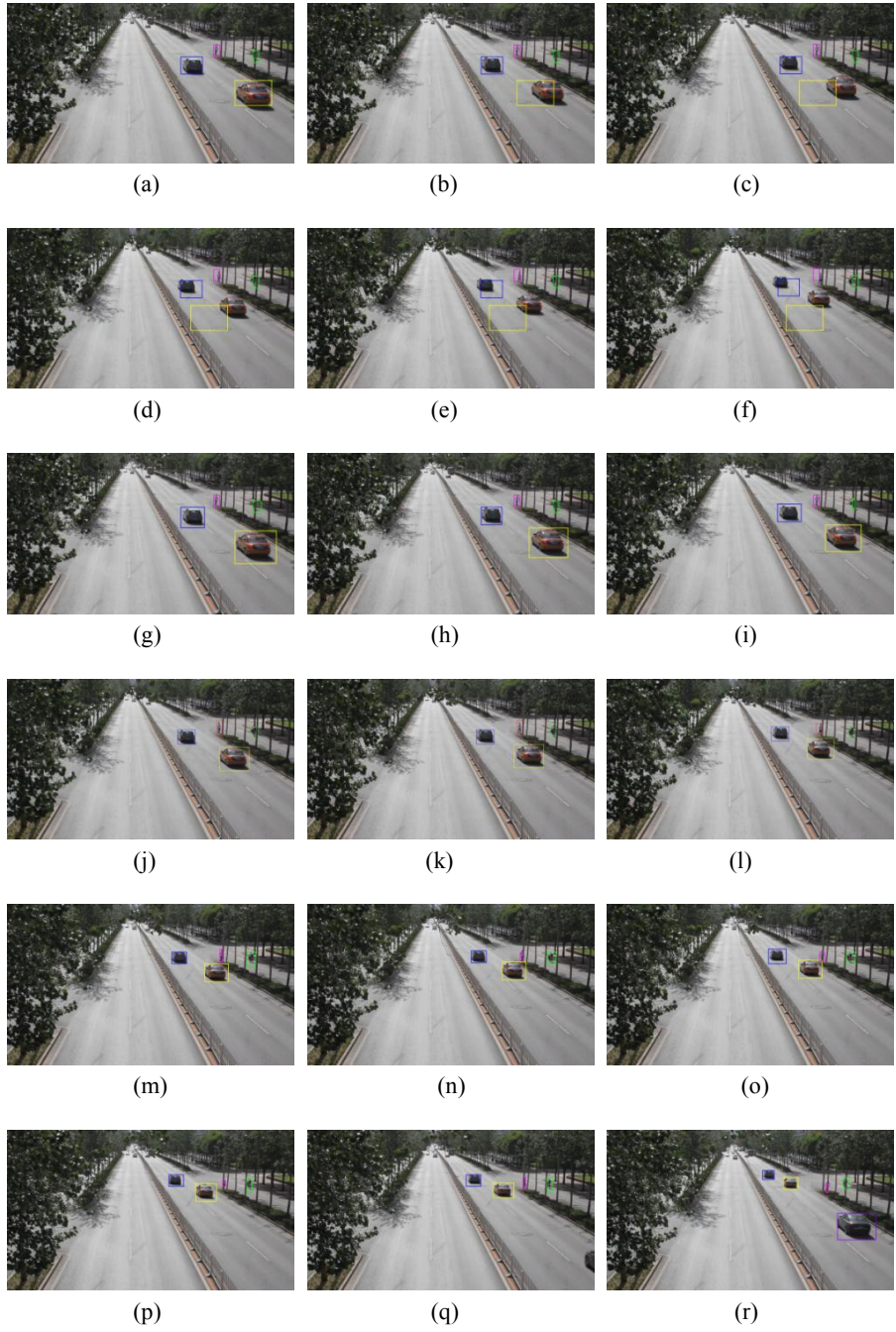

**Fig. 5.** Comparison of tracking effects

## Acknowlegements

This work was supported in part by the National Natural Science Foundation of P.R. China under Grant Nos. 61772574 and 62171482 and in part by the Basic and Applied Basic Research of Guangdong Province, P.R. China, under Grant No. 2021A1515011758.